# Solar radiation forecasting using ad-hoc time series preprocessing and neural networks [1]

Christophe Paoli[1], Cyril Voyant[1,2],
Marc Muselli[1], Marie-Laure Nivet[1]

[1] University of Corsica, CNRS UMR SPE 6134, (Vignola, Rte des Sanguinaires, 20000 Ajaccio | Campus Grimaldi, 20250 Corte), France
[2] Hospital of Castelluccio, Radiotherapy Unit, BP 85, 20177 Ajaccio, France

{christophe.paoli, cyril.voyant,
marc.muselli, marie-laure.nivet}@univ-corse.fr

**Abstract.** In this paper, we present an application of neural networks in the renewable energy domain. We have developed a methodology for the daily prediction of global solar radiation on a horizontal surface. We use an ad-hoc time series preprocessing and a Multi-Layer Perceptron (MLP) in order to predict solar radiation at daily horizon. First results are promising with nRMSE < 21% and RMSE < 998 Wh/m². Our optimized MLP presents prediction similar to or even better than conventional methods such as ARIMA techniques, Bayesian inference, Markov chains and k-Nearest-Neighbors approximators. Moreover we found that our data preprocessing approach can reduce significantly forecasting errors.

## 1 Introduction

An optimal use of the renewable energy needs its characterization and prediction in order to size detectors or to estimate the potential of power plants. In terms of prediction, electricity suppliers are interested in various horizons to estimate the fossil fuel saving and to manage and dispatch the power plants installed. Artificial intelligence techniques are becoming more and more popular in the renewable energy domain [1], [2] and particularly for the prediction of meteorological data such as solar radiation [3], [4] [5] [6]. Thereby many research works have shown the ability of Artificial Neural Networks (ANNs) to predict times series of meteorological data. In this study and according to electricity suppliers, we focus on the prediction of global solar irradiation on a horizontal plane for daily horizon. In this way, we have investigated time series forecasting which is a challenge in many fields. Because it has made tremendous progress in the past twenty years in terms of theory, algorithms and applications, we have chosen to study ANNs. Moreover, if we compare to conventional algorithms based on linear models, ANNs offer an attractive alternative by providing nonlinear parametric models. Through the proposed study, we will particularly look at the Multi-Layer Perceptron (MLP) network which has been the most used of ANN architectures in the renewable energy domain [1], [2]. The originality of our study is to add an ad-hoc time series preprocessing step before using neural networks. Indeed, as seen in [7] a data preprocessing including deseasonalization and detrending can improve ANN forecasting performance. As global solar radiation has a deterministic part, we want to take into account this specificity.

[1] A part of this research is founded by the Territorial Collectivity of Corsica.

The paper is organized as follow: after a brief presentation of the use of ANNs in time series prediction, the section 3 describes the physical phenomena we want to predict and introduces our ad-hoc time series preprocessing. Section 4 presents the neural network architecture we designed. Results are presented and discussed in section 5 where several conventional methods for estimation and modeling of the meteorological data are compared with our methodology. Section 6 concludes and suggests perspectives.

## 2 Time series prediction using neural networks

Time series prediction or time series forecasting takes an existing series of data $x_{t-n}$, .. , $x_{t-2}$, $x_{t-1}$, $x_t$ and forecasts the $x_{t+1}$, $x_{t+2}$ data values. The goal is to observe or model the existing data series to enable future unknown data values to be forecasted accurately. Thus a prediction $\hat{x}_t$ can be expressed as a function of the recent history of the time series, $\hat{x}_t = f(x_{t-1}, x_{t-2}, \ldots)$. ANNs represent a class of distinct mathematical models originally motivated by the information processing in biological neural networks, many of them applicable to forecasting tasks and modeling nonlinear functions *f*. The MLP [8], [9], [10] [11] represents a well researched non-recurrent ANN paradigm, which offers great flexibility in forecasting through flexibility in the number of input and output variables. Thus MLPs offer large degrees of freedom towards the forecasting model design. Figure 1 gives the basic architecture for a MLP application to time series forecasting. A fixed number *p* of past values is fed to the input layer of the MLP and the output is required to predict a future value of the time series. This method is often called the sliding window technique as the N-tuple input slides over the full training set.

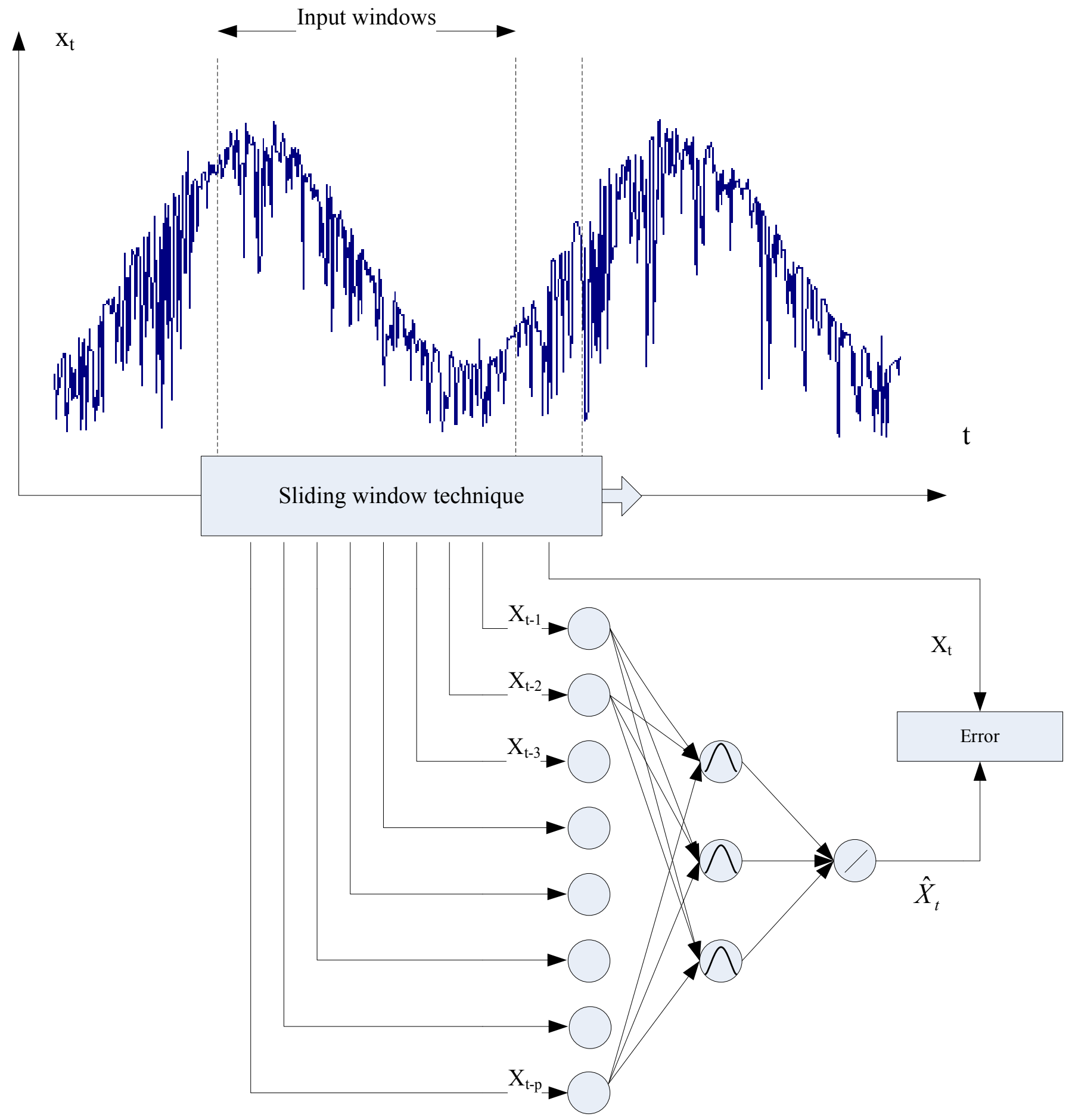

**Fig. 1.** MLP application to time series forecasting (bias nodes are not displayed).

Our study proposes to analyze the radiation time series (Wh.m-2) measured at the meteorological station of Ajaccio (METEO FRANCE, Corsica, France, 41°55'N, 8°44'E). The data representing the global solar radiation were measured on a daily basis from January 1971 to December 1989. Thus we have a $x_t$ time series to forecast for time t+1; that is at horizon 1. To achieve this we choose to use a gray box (or semi-physical) model [12] where time series prediction and modeling are mixed in order to profit by specificity. For time series prediction only past values are used to forecast the future values at a given horizon. In the case of modeling the different physical processes involved are taken into account in order to represent the variable however the horizon is.

In the remainder of this paper, we choose for the following naming convention: $X_t$ designate the time series, and $X_{d,y}$ is the modeling of the variables, where d is the day of the year y. In the next section, an explanation of the physical phenomenon is proposed, and then we describe our time series preprocessing.

## 3 Data analysis and preprocessing

There are two approaches that allow quantifying solar radiation: the "physical modeling" based on physical processes occurring in the atmosphere and influencing solar radiation [13], and the "statistic solar climatology" mainly based on time series analysis [13]. As already said we have chosen to combine these two methods in a gray box approach to improve the quality of prediction. In this work, we have used the physical phenomena in an attempt to overcome the seasonality of the resource.

### 3.1 Physical phenomenon

We can observe in Figure 2 that the global radiation consists of three types of radiation: direct radiation, diffuse radiation and ground-reflected radiation. The ground-reflected radiation does not concern us because we try to predict the radiation on a horizontal surface where the ground reflected radiation does not make no sense. For clear sky, ie without cloud cover, global radiation is relatively easy to model because it is primarily due to the distance from the sun sensor [14], [15], [16], [17], [18]. It is not the same, when there are clouds near the detectors. Indeed, these are mostly stochastic phenomena, which depend on the weather site.

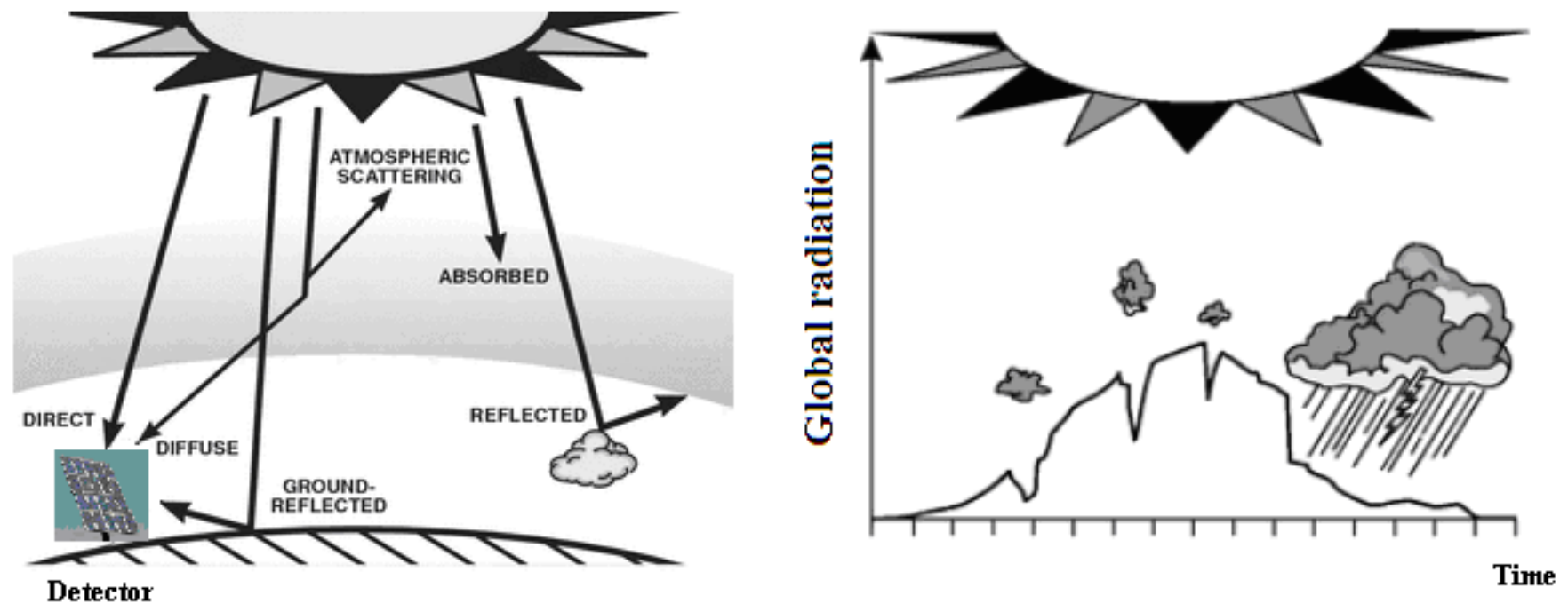

**Fig. 2.** Origin of the three types of radiation: direct radiation, diffuse radiation and ground-reflected radiation on a detector at ground level (left of the figure). Modification of the global irradiation profile accordingly to clouds cover (right of the figure).

The spectral analysis of the global radiation series highlights the high periodicity of the phenomenon (almost 365 days period). As proposed in [7], it appeared wise to make the time series stationary as much as possible. Deseasonalize and detrend a series allows to eliminate seasonal and trend components without changing the other information. In the present study, we have not considered the solar irradiation like a trend process but only like a seasonal process. The choice of the methodology used depends on the nature of the seasonality. As in our case the seasonality is very pronounced and repetitive, so very deterministic and not stochastic. Moreover, it is possible to physically quantify the components of our irradiation series. Like we have seen in figure 2, the global solar irradiance on horizontal plane depends on direct and diffuse radiation. This specificity allows us to apprehend the periodicity of the phenomenon, and to reduce the non-stationarity of the series.

In the next section, we propose to stationarize radiation data to overcome the deterministic component which is easily quantifiable. Thus we devote to the prediction of cloud cover included in the global radiation of the site.

### 3.2 Ad-hoc time series preprocessing

In fact, by dividing the series by daily extraterrestrial radiation [19], we can quantify the annual periodicity. It is the first step of our stationarization process. In first approximation, it is possible to quantify the deterministic component of global radiation by the extraterrestrial solar radiation alone ($H_0$). Thus we apply on the original series $X_{d,y}$ (where "d" is the day and "y" is the year) the ratio to trend method. This leads to a new series ($S_{d,y}$), known as series of index clarity:

$$S_t = S_{d,y} = \frac{x_{d,y}}{H_0^d} \quad \textbf{(1)}$$

After this step a new rigid seasonality is updated, we can lift it with the use of periodic seasonal factors [19]. This treatment aims to create a new distribution without periodicity. Although this pre treatment tends to stationarize the time series, a test of Fisher shows that seasonality was not optimal. According Bourbonnais [19], after using a ratio to trend method ($H_o$ in this case) to correct rigid seasonalities, we can use a ratio to moving average. This second ratio can be applied when there is no analytical expression of the trend. In our case, we find that $H_o$ led a new seasonality which is difficult to model. That's why we made a moving average ratio to overcome the seasonality. In the case of flexible seasonality, ie random in amplitude or period, the filtering techniques by successive moving averages are recommended.

$$y_{d,y} = \frac{S_{d,y}}{\frac{1}{2.m+1}.\sum_{i=-m}^{m} S_{d+i,y}} \quad \textbf{(2)}$$

In our case as 2m+1 = 365 days, we obtain m = 182. To complete the process, then we use the 365 seasonal factors ($y_d$). These are in fact coefficients which allow to overcome rigid seasonality by a moving average ratio described above. In order to not distort the series, we have considered that the total sum of the components of the series is the same before and after the report (final seasonal factors $y_d$* of the equation 5). The transition coefficients (N = 18, the number of years of history) and the average coefficients of the regular 365 days are given by the equations 3 and 4:

$$\overline{y_d} = \frac{1}{N}\left(\sum_{y=1}^{N} y_{d,y}\right). \quad \textbf{(3)}$$

$$\overline{\overline{y_d}} = \frac{1}{365}\left(\sum_{y=1}^{365} \overline{y_d}\right). \quad \textbf{(4)}$$

$$y_d^* = \frac{\overline{y_d}}{\overline{\overline{y_d}}}. \quad \textbf{(5)}$$

It follows a new series seasonally adjusted, that represents only the stochastic component of global radiation:

$$S_{d,y}^{corr} = \frac{S_{d,y}}{y_d^*}. \quad \textbf{(6)}$$

The following figure illustrates the effectiveness of pre-treatment on our time series. On the left we can see the spectrum of the original series. In the center we can see the result of a Fast Fourier Transform (FFT) on this series which highlights a peak of 365 days. And on the right we can see the result of a FFT on the same series, but after pre treatment. It is clear that the peak height at 365 days has disappeared.

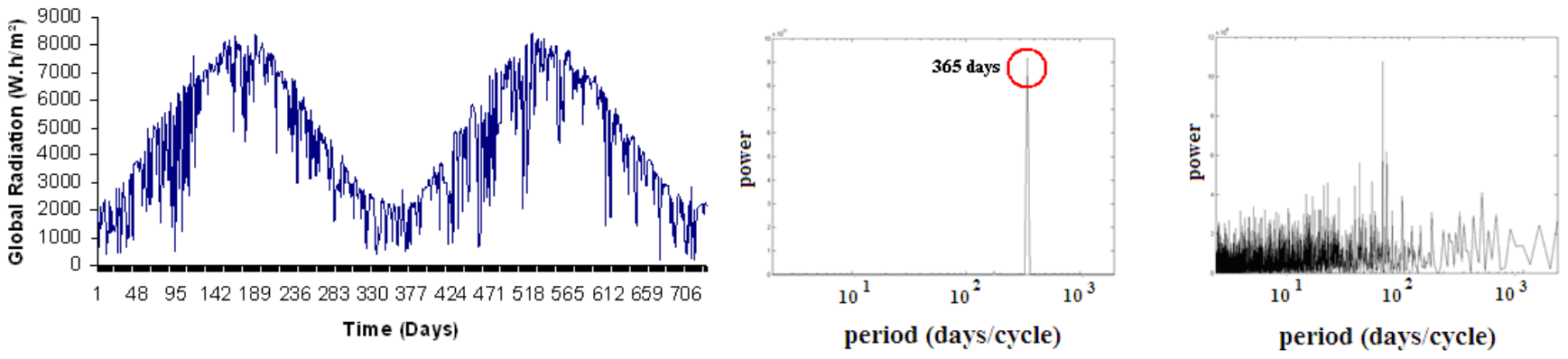

**Fig. 3.** Fast Fourier Transform of the original series.

After presenting the pre-treatment to be performed on our series and after verifying its effectiveness, we propose in the next section the architecture of RNA used.

## 4 Neural network architecture and design

The search for the ideal network structure is a complex and crucial task. We have adopted a feed forward Multi-Layer Perceptron (MLP), which is the most commonly used in the renewable energy domain [1], [2]. In order to determine the best network configuration, we have tried to study all the parameters available in this network architecture. To perform this optimization, we have considered the practice hypothesis that parameters are orthogonal. We have optimized parameters by considering each other constants.

As a result of this iterative process, the selected network has three neuron layers: input, hidden and output layer. There was no significant difference in the use of 1, 2 and 3 hidden layer architectures. One hidden layer was used in order to minimize the complexity of the proposed ANN model. The network has the following characteristics: eight neurons in the input layer which received as input the endogenous entries $S_{t-1}$,.. , $S_{t-8}$ normalized on {0,1}, three neurons on the hidden layer and one neuron on the output layer $\hat{S}_t$. Concerning the transfer functions and the training algorithm, the best results were obtained with the Gaussian (hidden layer) and linear (output layer) function and the Levenberg–Marquardt second-order algorithm.

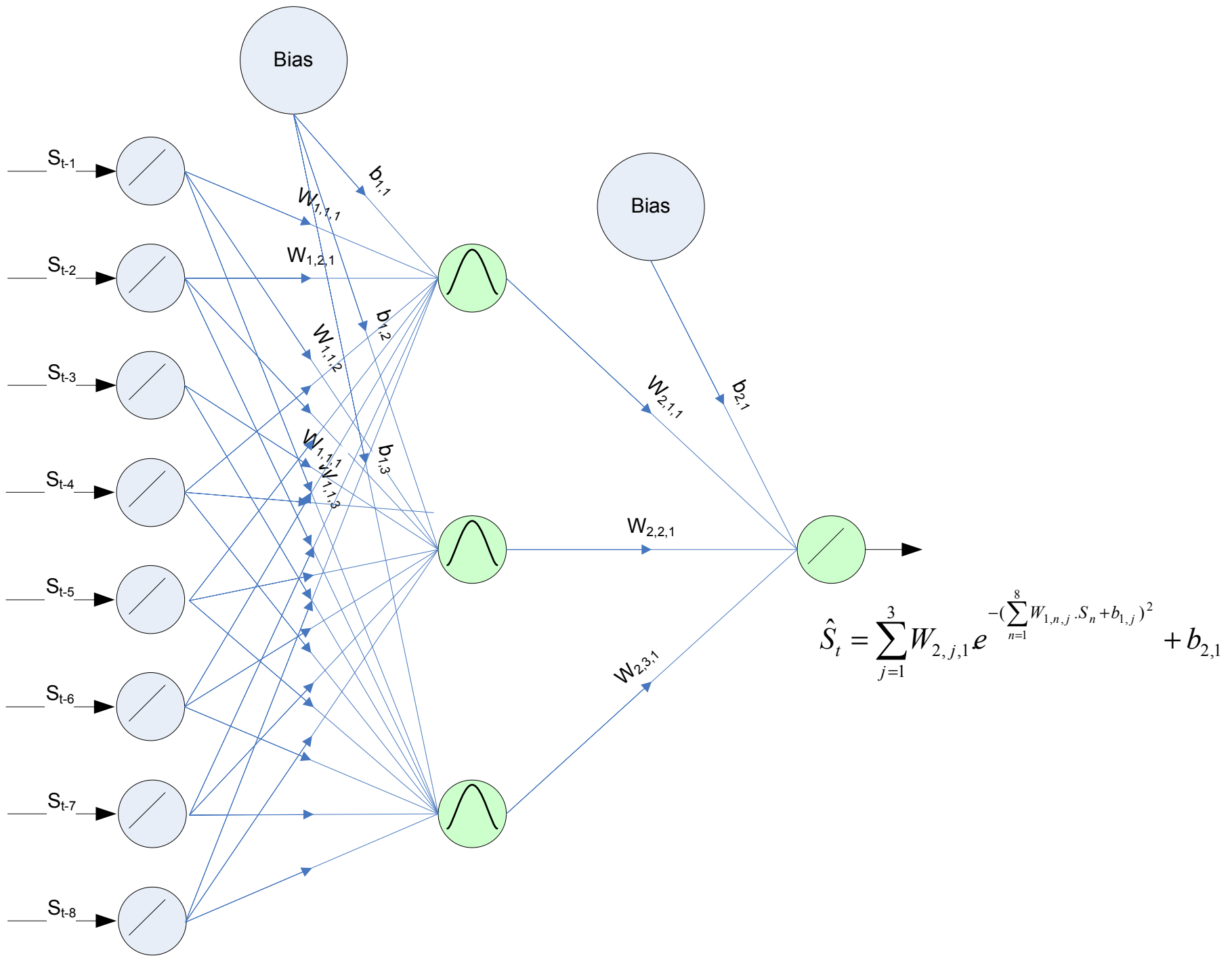

**Fig. 4.** Architecture of the optimized MLP.

We used the Matlab® software and its neural network toolbox to implement our network. The Matlab® training and testing data sets were set respectively to 80% and 20%. The early stopping technique was set to the maximum validation failure (the parameter max_fail = 5), the other training parameters were not considered to be significant. The learning has concerned the years 1971 to 1987 and the performance function was mean square error – MSE.

# 5 Results and discussions

Figure 5 summarizes the protocol that allowed us to conduct our experiences and validate our approach. A first treatment (step 1 of the figure 5) has allowed to clean the series of atypical points probably due to sensors maintenance or absences of measurement. We have replaced them by the average over the 19 years for the hours and the days corresponding to the problems.

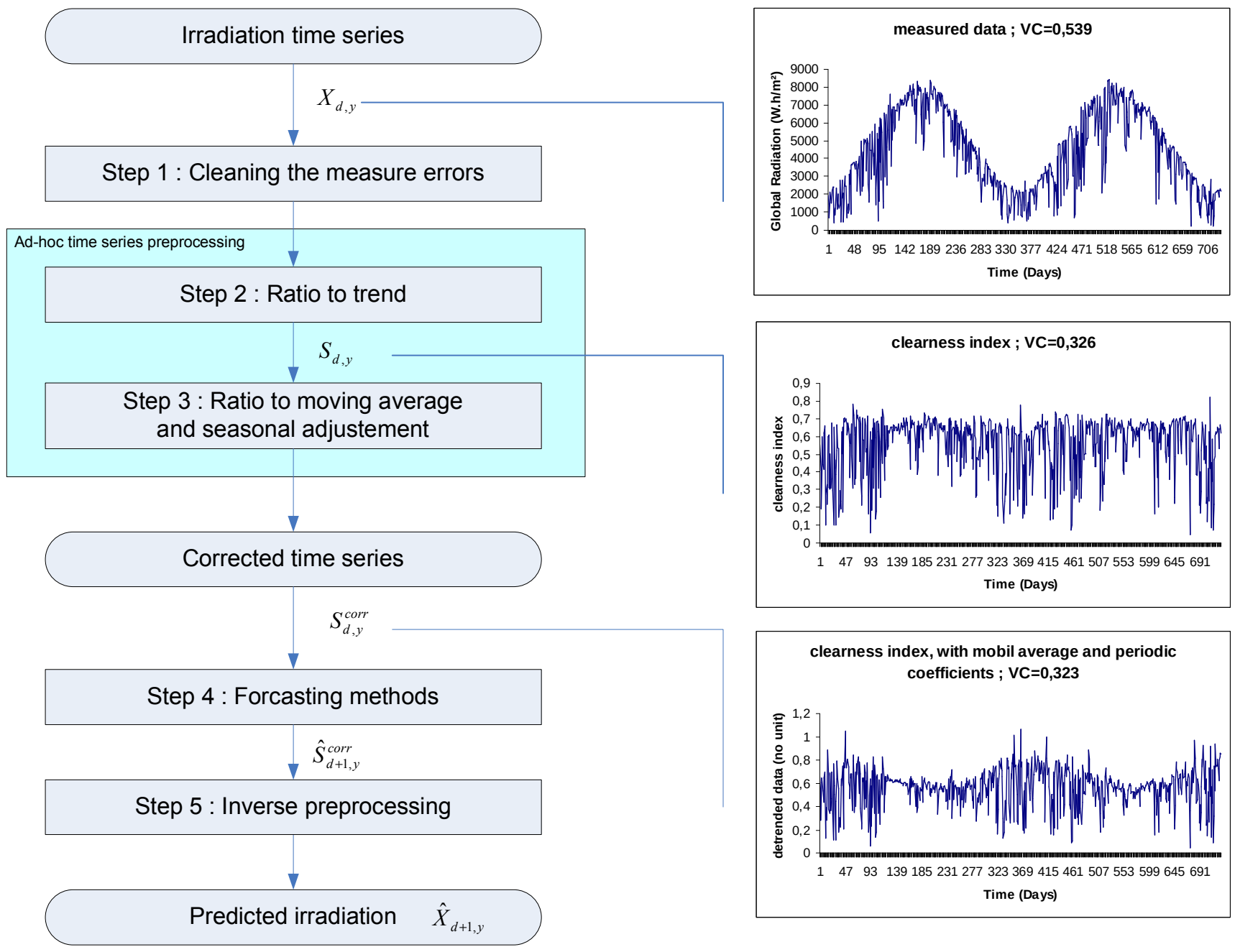

**Fig. 5.** Summarize of the protocol followed to obtain the predicted irradiation.

Steps 2 and 3 have been described in the previous section and lead to a series corrected. The step 4 was to compare classical forecasting methods outlined in the next section with our optimized MLP. Finally step 5 allows to reverse the preprocessing treatment and obtain the prediction of global irradiation.

## 5.1 The classical forecasting methods

In order the measure the effectiveness of our approach, we have decided to compare it with the following classical forecasting methods.

The ARIMA techniques are reference estimators in the prediction of global radiation field. It is a stochastic process coupling autoregressive component (AR) to a moving average component (MA). The interested reader may refer to [20], [21]. After several experiment we have obtained and decided to use an ARMA (2,2).

Bayesian inference is another classical technique. In this method evidences or observations are used to update or to newly infer the probability that a hypothesis may be true. The interested reader may refer to [22], [23]. In our experiences, we used Matlab software. We have identified that the prediction was better if we had 50 classes and an order of 3.

In forecasting domain, some authors have tried to use so-called Markov process, specifically the Markov chains. A Markov process is a stochastic process with the Markov property. Having the Markov property means that, given the present state, future states are independent of the past states. In other words, the description of the present state fully captures all the information that could influence the future evolution of the process. Future states will be reached through a probabilistic process instead of a deterministic one. The interested reader may refer to [24], [25]. In our case, we obtained 50 for the dimension of the transition matrix (number of class) and an order of 3 for the chain (determination of the prediction lag).

The k-nearest neighbors algorithm (k-NN) is a method for classifying objects based on closest training examples in the feature space. k-NN is a type of instance-based learning, or lazy learning where the function is only approximated locally and all computation is deferred until classification. It can also be used for regression. Unlike previous ones this tool does not use a learning base. The method consists in looking into the history of the series for the case the most resembling to the present case. The interested reader may refer to [26]. In our study we choose a k equal to 10.

The following section presents the results obtained.

### 5.2 Results

To determine whether our network was really interesting in terms of daily prediction of solar radiation, we compared its performances with the forecasting results obtained with a naive predictor (average over 18 years of the day considered), order 3 Markov chains, order 3 Bayesian inferences, an order 10 k-NN, an order 8 AR without preprocessing, an ARMA(2,2) with preprocessing. Table 1 presents results we have obtained in the case of an annual error for daily prediction of global solar radiation.

**Table 1.** Annual error for all prediction methods, forecasting years 1988 and 1989, 1 day horizon

| | **nRMSE** |
|---|---|
| Naïve predictor | 26 % |
| Markov Chain (order 3) | 25,11 % |
| Bayes (order 3) | 25,16 % |
| k-NN (order 10) | 25,20 % |
| AR(8) without preprocessing | 21,18 % |
| ARMA(2,2) with preprocessing | 20,31 % |
| ANN[8,3,1] without preprocessing | 20,97 % |
| ANN[8,3,1] with preprocessing | 20,17 % |

We highlight that the predictors other than ARMA and ANN give the same results, slightly higher than those obtained with a naive predictor. Even without preprocessing, ARMA and ANN are the best predictors. The preprocessing improves the quality of prediction and allows access to the 20% error.

The predicted results for each combination were compared statistically using three parameters: the Root Mean Square Error (RMSE), the normalized RMSE (nRMSE), and the Mean Bias Error (MBE):

$$RMSE = \sqrt{\frac{1}{N}\sum_{i+1}^{N}(C_i - M_i)^2} \cdot \quad \mathbf{(6)}$$

$$nRMSE = \frac{\sqrt{\frac{1}{N}\sum_{i+1}^{N}(C_i - M_i)^2}}{\sqrt{\frac{1}{N}\sum_{i=1}^{N}(M_i)^2}} \cdot \quad \mathbf{(7)}$$

$$MBE = \frac{1}{N}\sum_{i+1}^{N}(C_i - M_i) \cdot \quad \mathbf{(8)}$$

$C_i$ and $M_i$ are respectively the $i_{th}$ calculated and measured values and N is the total number of observation. Table 2 details in the MLP case, the annual prediction errors obtained for the years 1988 and 1989.

**Table 2.** Annual prediction error for the years 1988 and 1989 with our MLP

| | **Arithmetic Mean** | **95% confidence interval** |
|---|---|---|
| nRMSE | 20,2 % | 0,1 % |
| RMSE | 997,971 | 6,333 |
| MBE | -104,239 | 28,794 |
| $R^2$ | 0,801 | 0,002 |
| Monthly average error | 3,9 % | 0,4 % |

The confidence interval is calculated after 10 simulations. Given the small size of the confidence intervals, we can say that there are very few local minimums. The monthly average error represents the error for the value of the irradiation. We obtain that the combination of the prediction of global radiation received after 1 month is different from an average of 4% of the aggregate measured. The negative MBE means that we underestimate the solar potential on averaged over the year. Since we have an atypical day of low irradiation then there is a tendency to overestimate. The determination coefficient $R^2$ is greater than 0,8.

Figure 6 shows the errors of prediction and distinguishes the seasons for the years 1988 and 1989. As we can see best results; i.e. less important error, in term of forecast are obtained in summer. These results can be used for example by energy managers who need to avoid using hydraulic power plants in dry season.

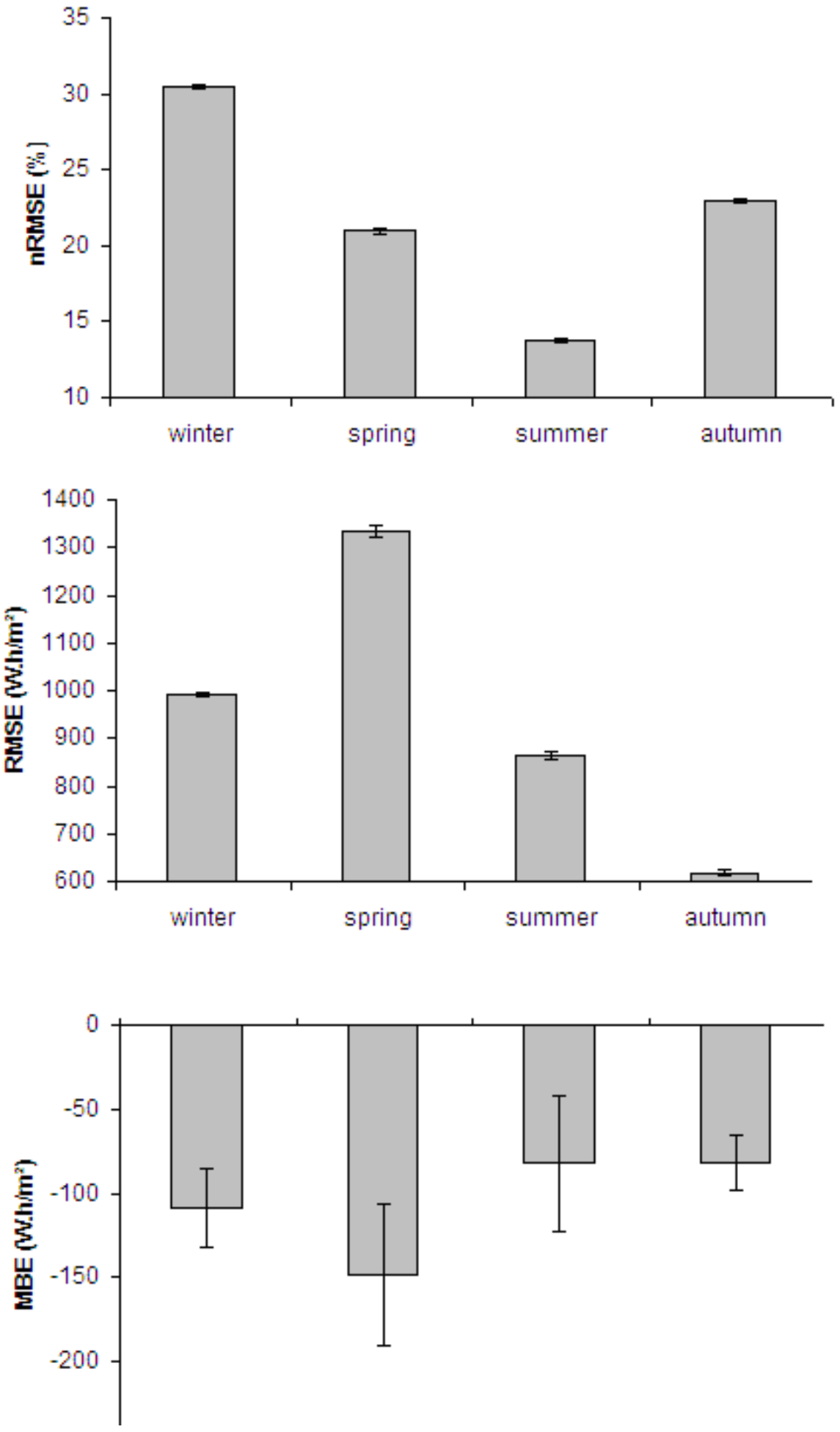

**Fig. 6.** Seasonal errors for the daily prediction of the years 1988 and 1989 (mean with 95% confidence interval).

There should be a compromise between RMSE and nRMSE. The nRMSE are useful for comparison and optimization. But for the interpretation of energy, we must look at the RMSE. The spring season is the most difficult to predict. The absolute error is consistent. However, we find that in summer the error does not exceed 900 Wh / m², while the irradiation is important. MBE are found negative, which indicates an underestimation. The MBE is not significantly different from one season to another. Thus we will always have the same prediction error, whatever the season.

Finally, Figure 7 compares the real data of solar radiation with the results obtained with our MLP with preprocessing. The error of prediction is also drawn.

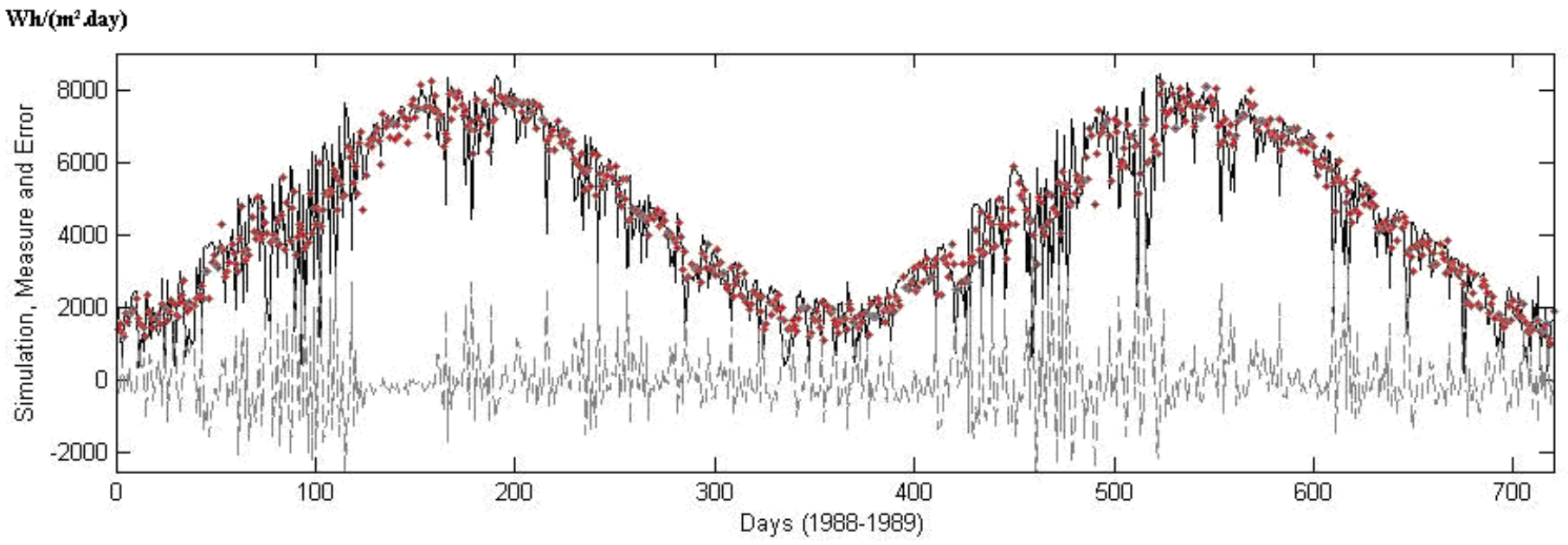

**Fig. 7.** Comparison of the real data of solar radiation with the results obtained with our MLP with preprocessing and error of prediction. Dashed line is the error; solid line is the real data of solar radiation and red points are the prediction.

There is an increase in errors at the beginning of the cycle; this corresponds to the spring when the cloud disturbances are very important. We can see it is very difficult to predict the radiation cause of very noisy variable. Figure 8 shows the correlation between experimental and simulated global irradiation.

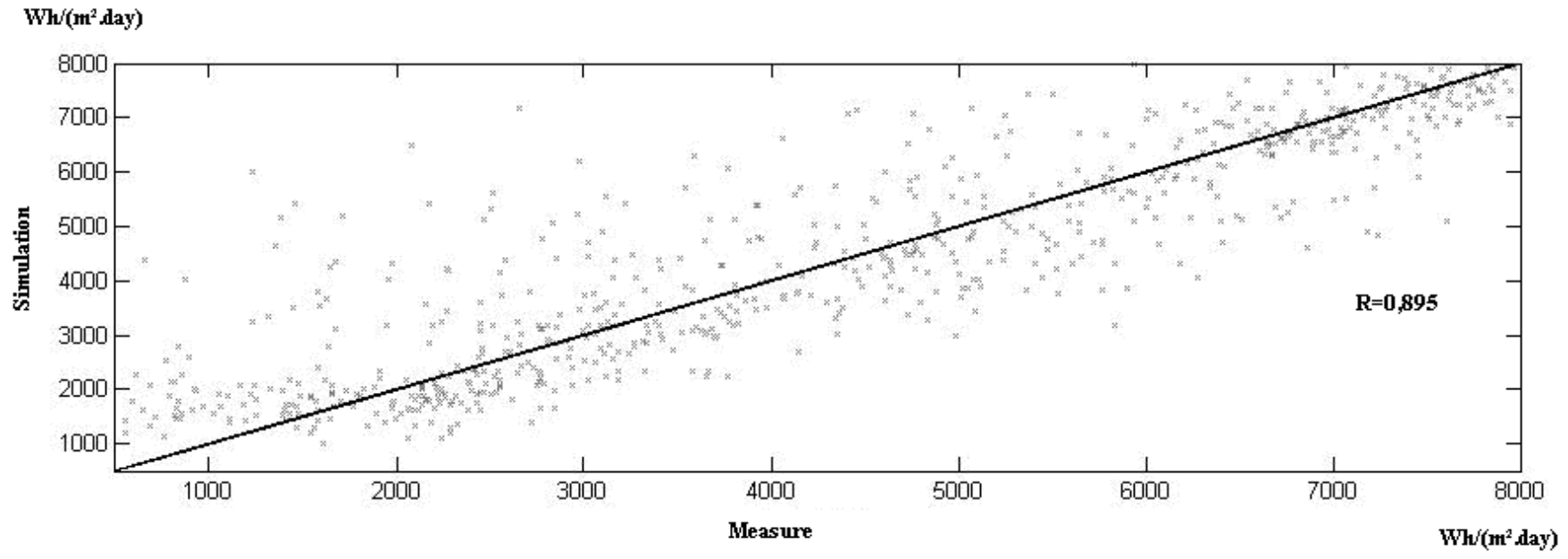

**Fig. 8.** Correlation between experimental and simulated global irradiation.

We systematically overestimate the days when the irradiation is minimal (winter). The points that lie at the very top of the line y = x shows that it is very difficult to predict the days when the irradiation had to be theoretically important. We would undoubtedly have improved the results optimizing an ANN by season, but it would complicate the procedure, and tend to decrease the robustness of the procedure. The next section concludes this paper and suggests prospects.

## 6 Conclusion and perspectives

This paper has developed an ANN predictor approach to determine global irradiation at daily horizon in order to help electrical managers. We have used an ad-hoc time series preprocessing and a time series prediction designed MLP. Although the location was very specific, with the proximity to the sea and the mountain that can greatly affect nebulosity, we have obtained relevant results. Seasonal RMSE are less than 998 Wh/m-2 (nRMSE < 21%). ANN processes presents a great interest compared to classical stochastic predictor like ARIMA. Moreover we found that our data preprocessing approach reduce significantly forecasting errors.

The next step of our work will be to validate our predictor on real photovoltaic system. It was recently installed in our laboratory and we are awaiting data. In the future, it seems important to study shorter time horizons. As matter of fact, electrical managers are also interested to horizons that can range from ½ hour to several hours: from 3 hours to 24 hours. Thus others ANN architecture types have to be studied: recurrent ANNs, adaptative ANNs, etc. An ongoing study will be based on the implementation of exogenous variables on the input neurons like METARs data (pressure gradient, temperature, etc.). Determining the relevant variables could be done by the random probe method [27].

In the long-term, it would be also very interesting to study a network trained on a city, and used on other site with equivalent geographical feature, and maybe combine both ANN and Geographic Information Systems (GIS) approaches.